\documentclass[10pt,twocolumn,letterpaper]{article}

\usepackage{wacv}
\usepackage{times}
\usepackage{epsfig}
\usepackage{graphicx}
\usepackage{amsmath}
\usepackage{amssymb}
\usepackage{booktabs}
\usepackage{flushend}

\def\wacvPaperID{1649} 

\wacvalgorithmstrack   

\wacvfinalcopy 

\ifwacvfinal
\usepackage[breaklinks=true,bookmarks=false]{hyperref}
\else
\usepackage[pagebackref=true,breaklinks=true,colorlinks,bookmarks=false]{hyperref}
\fi

\usepackage{lipsum} 
\usepackage{bbm}
\usepackage{scalerel}
\usepackage{multirow}
\usepackage{float}
\usepackage[normalem]{ulem}
\usepackage{array}

\usepackage[capitalize]{cleveref}
\crefname{section}{Sec.}{Secs.}
\Crefname{section}{Section}{Sections}
\Crefname{table}{Table}{Tables}
\crefname{table}{Tab.}{Tabs.}

\def\R{\mathbb{R}}
\DeclareMathOperator*{\OR}{\scalerel*{+}{\sum}}
\newcolumntype{C}[1]{>{\centering\arraybackslash}p{#1}}

\newcommand\blfootnote[1]{%
  \begin{NoHyper}
  \begingroup
  \renewcommand\thefootnote{}\footnote{#1}%
  \addtocounter{footnote}{-1}%
  \endgroup
  \end{NoHyper}
}

\begin{document}

\title{On Utilizing Relationships for Transferable Few-Shot Fine-Grained Object Detection}

\author{Ambar Pal\\
Johns Hopkins University\\
{\tt\small ambar@jhu.edu}
\and
Arnau Ramisa, Amit Kumar K C, René Vidal \\
Amazon Visual Search \& AR\\
{\tt\small \{aramisay, amitkrkc, rvidal\}@amazon.com}
}

\maketitle
\thispagestyle{empty}

\begin{abstract}
State-of-the-art object detectors are fast and accurate, but they require a large amount of well annotated training data to obtain good performance. However, obtaining a large amount of training annotations specific to a particular task \ie, fine-grained annotations, is costly in practice. In contrast, obtaining common-sense relationships from text,  \eg “a table-lamp is a lamp that sits on top of a table”, is much easier. Additionally, common-sense relationships like ``on-top-of'' are easy to annotate in a task-agnostic fashion. In this paper, we propose a probabilistic model that uses such relational knowledge to transform an off-the-shelf detector of coarse object categories (\eg “table”, “lamp”) into a detector of fine-grained categories (\eg “table-lamp”). We demonstrate that our method, \textsc{RelDetect}
achieves performance competitive to finetuning based state-of-the-art object detector baselines when an extremely low amount of fine-grained annotations is available ($0.2\%$ of entire dataset). We also demonstrate that \textsc{RelDetect} is able to utilize the inherent transferability of relationship information to obtain a better performance ($+5$ mAP points) than the above baselines on an unseen dataset (zero-shot transfer). In summary, we demonstrate the power of using relationships for object detection on datasets where fine-grained object categories can be linked to coarse-grained categories via suitable relationships.
\end{abstract}

\section{Introduction}
\begin{figure*}[t!]
\centering
\includegraphics[width=\textwidth]{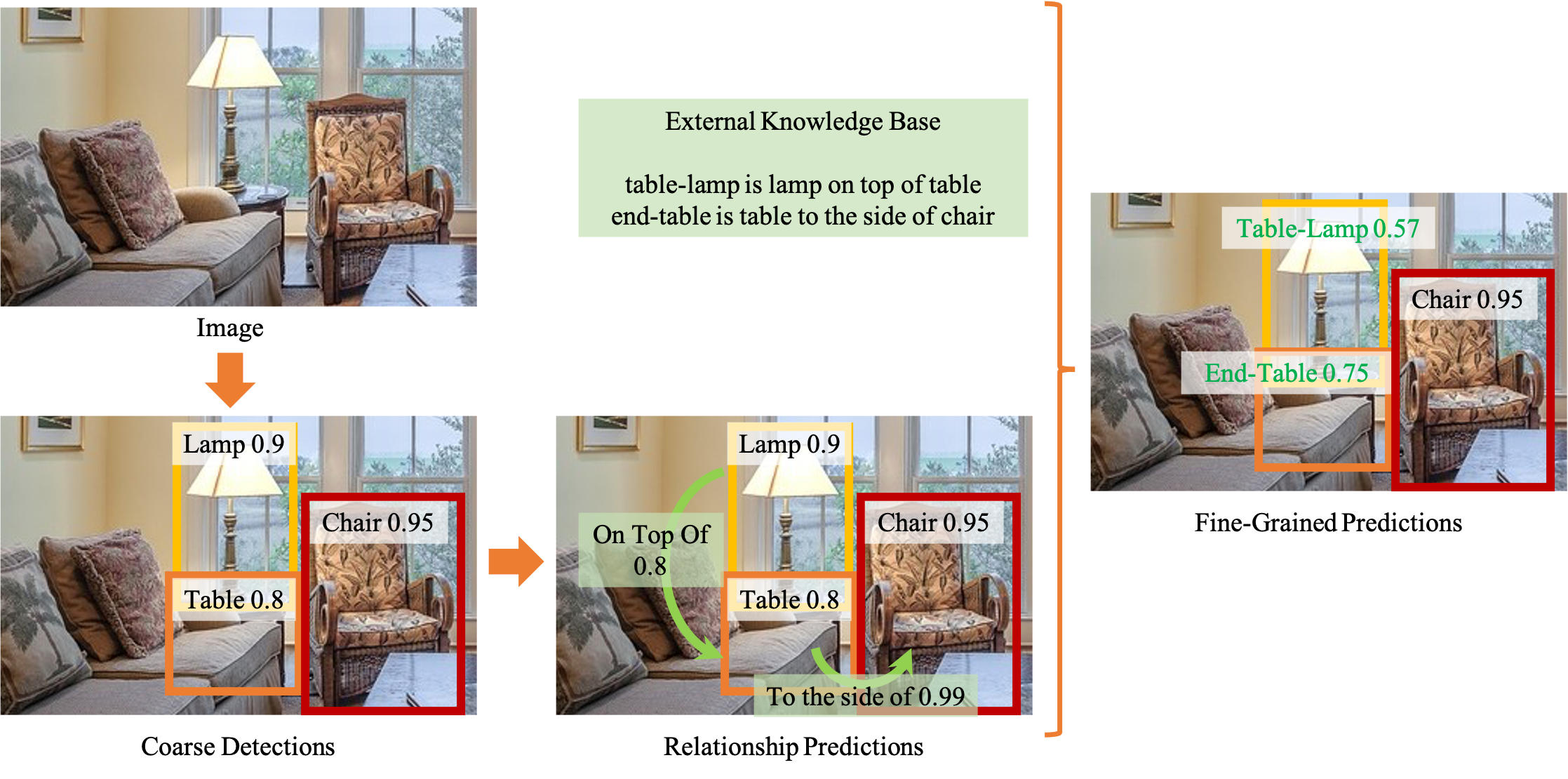}
\caption{Overview of the proposed method. Given an image, we firstly use an object detector trained on base classes to produce detections for \texttt{lamp}, \texttt{table}, \texttt{chair} and their corresponding class probabilities. A relationship predictor then takes in these detections as input and produces possible relations between various objects in the image. Finally our model looks into a relational knowledge base (details in \cref{sec:kb}) to deduce the fine-grained categories \texttt{table-lamp} and \texttt{coffee-table}.}
  \label{fig:teaser}
\end{figure*}
\label{sec:intro}

\blfootnote{This work was done while AP was a research intern at Amazon. AP is with the Department of Computer Science, and the Mathematical Institute for Data Science at Johns Hopkins University.} Object detection is an essential component of many computer vision applications. Accordingly, steady progress over the past decade has led to modern object detectors that are both fast and accurate \eg \cite{ren2016faster,redmon2018yolov3}. However, these state-of-the-art object detectors require large amounts of training data in order to obtain good detection performance. For instance, while off-the-shelf object detectors pre-trained on large public datasets like MS-COCO \cite{lin2014microsoft} can detect coarse-grained classes like ``table'', ``lamp'', and ``chair'' very well, they struggle at detecting and differentiating among derivative fine-grained classes like ``coffee-table'', ``floor-lamp'' and ``table-lamp''. 
Obtaining training data for such fine-grained classes is very expensive and when obtained, training state-of-the-art detectors on such large amounts of task-specific data requires significant computational resources. All these factors create an important practical bottleneck in the applicability of state-of-the-art detectors to specialized tasks.

On the other hand, common sense \emph{relational knowledge} that links coarse-grained to fine-grained classes is readily available from text, \eg ``a table-lamp is a lamp that sits on top of a table'', ``a coffee-table is a table in front of a sofa or chair arrangement''. Moreover, detecting and labelling relationships between objects in an image, such as ``on top of'' and ``in front of'', has been studied widely under the name of scene-graph generation \cite{yang2018graph,zellers2018neural,tang2020unbiased,suhail2021energy}. One might immediately wonder whether annotating relationships 
requires an even higher annotation effort. While the number of general spatial relationships between objects in an image can be large (quadratic in the number of object instances), general relationships are often not very discriminative for the task of fine-grained classification. Discriminative relationships often occur between a few pairs of objects, and are much easier to label. Additionally, such discriminative relationships can often be derived from fine-grained labels using automatic rules (as we shall see in \cref{sec:fsod}), eliminating the need for an expert annotator. 
Thus annotation of discriminative relationships is a cheaper and much less time-consuming task than fine-grained object bounding box annotation. 

Such advances in object detection and relationship prediction motivate us towards the central question that we aim to investigate in this work: \emph{Can we combine off-the-shelf coarse-grained object detection networks and relationship prediction networks to obtain a
fine-grained object detector that achieves a performance comparable to that of the state of the art while requiring much fewer training labels?}

In this work, we propose a principled approach to tackle the above question which builds upon observations of humans solving the fine-grained annotation task: Firstly, they identify the coarse-grained labels in the image, say ``lamp'' and ``table''. Secondly, they identify the relationships between the coarse-grained objects, say ``on top of''. 
Finally, they use the relationship information available to label the fine-grained object, say ``table-lamp''. We propose 
a probabilistic model that factors into products corresponding to the steps of the human reasoning process: identify the coarse-grained objects \ie,
\begin{align*}p({\rm Object Bounding Boxes, Object Coarse Labels | Image}),\end{align*} identify relationships given coarse-grained objects \ie, \begin{align*}p({\rm Relationship Labels | Objects}),\end{align*} and finally identify fine-grained objects given relationships and coarse-grained objects \ie, \begin{align*}p({\rm Object FineGrained Labels | Relationships, Objects}).\end{align*} We evaluate the effectiveness of our method for two tasks: Few-Shot Learning for a source domain and Zero-Shot transfer learning from a source domain to a target domain. We find that exploiting relationships significantly reduces the need for fine-grained levels and leads to more transferable representations.

In summary, we make the following contributions:
\begin{enumerate}
\item We propose a probabilistic model we call \textsc{RelDetect} for detection of fine-grained object classes. The model has three main components. First, an object detection model for coarse-grained classes. Second, a detector of spatial relationships among detected objects in the scene. Third, a relational knowledge model linking the coarse classes to the fine-grained classes via the detected spatial relationships to obtain a fine-grained class for each detected object.

\item We demonstrate the effectiveness of our approach for few-shot object detection on the DeepRooms dataset~\cite{gadde2021detail} suited for furniture detection on images of indoor home scenes. We evaluate our method quantitatively using the COCO mAP@50 metric \cite{lin2014microsoft} 
commonly used in object detection. We demonstrate that our method \textsc{RelDetect} is able to obtain few-shot performance competitive with state-of-the-art baselines when training with $0.2\%$ of the available fine-grained labels. We further evaluate qualitatively with examples of success and failure cases. 

\item We demonstrate the effectiveness of our approach for zero-shot domain transfer from DeepRooms to LivingRoom, 
a dataset having the same classes as DeepRooms but a different data distribution. Our experiments show that learning using relationships leads to transferrable representations that perform better than fine-tuning baselines. 

\end{enumerate}

\section{Related Work}

\textbf{Object Detection} has improved significantly since the advent of deep learning, and nowadays it is often used as a first step in many computer vision processing pipelines. Object detectors are typically divided into one-stage~\cite{redmon2018yolov3} and two-stage~\cite{ren2016faster}, with the former usually regarded as being faster at inference time, and the later being easier to train and more accurate. Although Convolutional Networks have dominated the state-of-the-art for object detection for a long time, new transformer-based models are starting to challenge their dominance~\cite{xu2021end}.

Much of the object detection literature is built around the MSCOCO~\cite{lin2014microsoft} dataset, with comparatively few works tackling related problems like object detection under domain shift~\cite{oza2021unsupervised} or fine-grained object detection. In Yang et al.~\cite{yang2019detecting}, the authors use existing image-level annotations for fine-grained categories to augment the training of a detector on a coarse-level bounding box dataset via visual correlations, producing a fine-grained detector. In contrast, in our work we do not assume a pre-existing fine-grained classification dataset, but only common-sense knowledge on the spatial relationships between the coarse-level and fine-grained object categories, as well as a generic relationship classifier.


 Few-shot object detection can also potentially reduce the high cost of labelling many fine-grained categories, by learning to detect new categories given only a few examples. 
 Wang et al.~\cite{wang2020frustratingly} propose to fine-tune only the last layer of an existing detector on rare clases, to drastically reduce the amount of data needed to fit the parameters, while leveraging the intermediate representations learned in the baseline training. Sohn et al.~\cite{sohn2020simple} uses pseudo-labeled data and strong data augmentations to obtain improved performance in low labelled data regimes.

\textbf{Scene Graph Generation (SGG)} focuses on determining relationships between objects and their attributes in an image as a basis for higher level understanding of the visual scene. Yang et al.~\cite{yang2018graph} first determine relevant relations between object proposals using a Relation Proposal Network, and then they update the resulting graph using an attention Graph Convolutional Network to exploit correlations between the object and relationships to improve prediction accuracy. Zellers et al.~\cite{zellers2018neural} analyze motifs (regularly appearing substructures in scene graphs) and determine a frequency prior that results on improved performance. Tang et al.~\cite{tang2020unbiased} use causal and counterfactual reasoning to remove bias induced by the suboptimal annotation of most SGG dataset, where certain simple relationship labels are heavily overused with respect to more informative ones. Suhail et al.~\cite{suhail2021energy} propose an energy-based loss for scene graph generation that takes into account the structure of the scene, instead of treating the objects and relationships as independent entities. 
Recently, Han et al.~\cite{han2021image} surveyed the existing literature and published a benchmark to compare SGG methods. 

\section{Probabilistic Model for Relationship-Based Fine-Grained Detection}
\textbf{Problem Statement} Assume we are given an off-the-shelf object detector trained to detect $\tilde K$ coarse classes and a relational knowledge base $B$ that links coarse classes to $K$ fine-grained classes via spatial relationships of $L$ different types. For example, the knowledge base entry ``a table-lamp is a lamp that sits on top of a table'' links the coarse classes ``lamp'' and ``table'' to the fine-grained class ``table-lamp'' via the relationship ``on top of''. Assume further we are given a small dataset of images of coarse-grained class objects annotated with a few relevant spatial relations. Our task is to build a detector to accurately localize and classify objects belonging to fine-grained derived classes with little to no training examples for the fine-grained classes.



\textbf{Notation} Let $I$ denote the input image and let $\{S_i\}_{i=1}^N$ denote the spatial bounding box descriptions for $N$ detected bounding boxes, \eg their location and size. Let $\tilde{C_i} \in \{1, \ldots \tilde K\}$ and $C_i \in \{1, \ldots K\}$ denote, respectively, the coarse and fine-grained classes for bounding box $i$, and let $R_{i, j} \in \{1, 2, \ldots, L + 1\}$ denote the 
relation between bounding boxes $i$ and $j$. The relation $L + 1$ denotes \emph{no-relation}. Finally, given a bounding box $i$, the set ${\rm NB}_i$ denotes the spatial neighbors of $i$.

\textbf{Probabilistic Model}
Given image $I$, our joint probabilistic model $p(C, R, \tilde C, S | I)$ can be factored as 
\begin{equation}
p(C | R, \tilde C, S, I) p(R | \tilde C, S, I) p(\tilde C | S, I) p(S | I) \label{eq:factorize}.
\end{equation}%
In what follows, we define each term in \cref{eq:factorize}, and make some simplifying assumptions to make the corresponding part of our model tractable.

\textbf{Region Proposal} The term $p(S | I)$ can be seen as a region proposal network \texttt{RPN}, which takes an image as input and produces a collection of bounding boxes $S=\{S_i\}_{i=1}^N$. The term $p(\tilde C | S, I)$ can be seen as an object classification head \texttt{OBJ} which predicts the coarse-grained base class of the objects given the image and the bounding boxes. Hence, the factor $p(S | I) p(\tilde C | S, I)$ represents an off-the-shelf object detector for coarse classes. \cref{sec:rpn} details the specific implementation for \texttt{RPN} and \texttt{OBJ} we use. 


\textbf{Relationship Prediction} The term $p(R | \tilde C, S, I)$ predicts all relationships given all the coarse object classes and their corresponding bounding boxes. In order to enable make the inference more tractable, we make the following simplifying assumption:
\begin{enumerate}
\item[A1.] \emph{The relationships are conditionally independent given the coarse classes, the image and the spatial coordinates of the bounding boxes}. That is, $\{R_{i, j}\}_{i,j=1}^{N}$ are conditionally independent given $I, S$. Intuitively, this assumption says that the relationships do not affect each other once we know the bounding boxes, their image features, and their coarse classes. 
\end{enumerate}
We simplify $p(R | \tilde C, S, I))$ using assumption A1 as
\begin{align}
p(R | \tilde C, S, I) = \prod_{i, j} p (R_{i, j} | \tilde C, S, I) \label{eq:relsplit}.
\end{align}

Each of the terms $p(R_{i, j} | \tilde C, S, I)$ can be seen as a relationship prediction network \texttt{REL} that takes in object bounding boxes, their image features and the coarse class predictions as input, and produces the relationship between boxes $i$ and $j$ as output. We utilize a state-of-the-art network architecture from the Scene Graph Generation literature for \texttt{REL} and detail the resultant implementation in \cref{sec:relpn}. 

\paragraph{Knowledge Base Query} The term $p(C | R, \tilde C, S, I)$ finally predicts all the fine-grained classes given all the relationships, coarse object classes and their bounding boxes. Again, we can make inference more tractable by making the following simplifying assumption:
\begin{enumerate}
\item[A2.] \emph{The fine-grained class of a bounding box is conditionally independent of (a) the coarse classes of bounding boxes not in its neighborhood, 
(b) its relationships with boxes not in its neighborhood, and (c) the relationships among other boxes, given the coarse classes and relationships in its neighborhood %
\footnote{
This definition is generic, but for our experiments we take the neighborhood ${\rm NB}_i$ to be the set of bounding boxes $j$ whose centers are at most a fixed distance $\tau$ from the center of box $i$.}
.} That is, $C_i$ is conditionally independent of $\tilde C_j$ and $R_{ij}$ given $I, S$ when $j \not \in {\rm NB}_i$, and $C_i$ is conditionally independent of $R_{j, k}$ given $I, S$ when $j\neq i$ and $k\neq i$. Intuitively, this assumption says that objects very far away in the scene do not affect each other once we know the objects in the neighborhood and the relationships among them. 
\end{enumerate}
Using assumption A2, we obtain the simplification 
\begin{equation}
p(C | R, \tilde C, S, I)  = \prod_i p(C_i | R_{i, {\rm NB}_i}, \tilde C_{{\rm NB}_i}) \label{eq:novelsplit},
\end{equation}
where each of the factors $p(C_i | R_{i, {\rm NB}_i}, \tilde C_{{\rm NB}_i})$ can be seen as querying the relational knowledge base $B$. As an example, let $C_1$ be the bounding box for a ``lamp'' and its neighborhood ${\rm NB}_1 = \{1, 2\}$ be the objects spatially close to this lamp in the scene. If the knowledge base contains ``table-lamp is lamp on top of table'', then the conditional probability $p(C_1 = \textsc{table-lamp} | C_1 = \textsc{lamp}, R_{1, 2} = \textsc{on top of}, C_2 = \textsc{table})$ should be $1$. This can be implemented by a rule-based model detailed in \cref{sec:kb}. 

\begin{figure}[t]
\centering
\includegraphics[width=\linewidth]{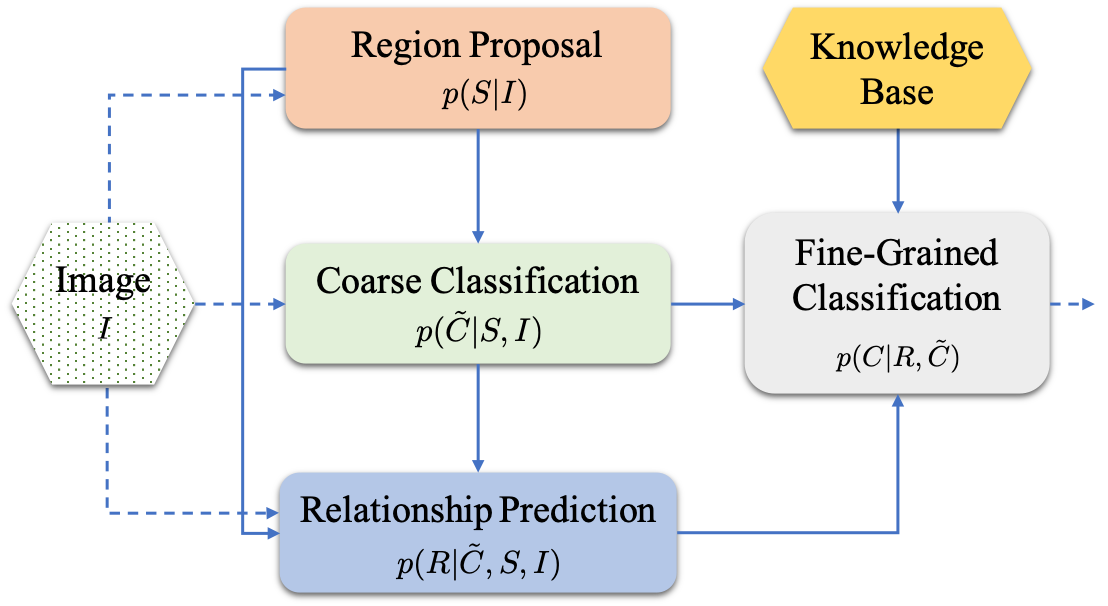}
\caption{Different components of our proposed model. Given an image $I$, the region proposal model is used to produce spatial coordinates $S$ of the object bounding boxes. This is then used in the coarse classification model to produce the coarse classes $\tilde C$. Then, the relationship prediction model takes the coarse classes and bounding boxes as inputs to produce the relationships $R$. Finally, the fine-grained classes $C$ are obtained by utilizing the knowledge base $B$.}
  \label{fig:overview}
\end{figure}

Finally, we put together \cref{eq:relsplit,eq:novelsplit,eq:factorize} to obtain the full factorization:
\begin{align}
p(C, R, \tilde C, S | I) &= p(S | I) p(\tilde C | S, I) \prod_{i, j} p (R_{i, j} | \tilde C, S, I) \nonumber \\
&\quad  \prod_i p(C_i | R_{i, {\rm NB}_i}, \tilde C_{{\rm NB}_i}) \label{eq:finalfactorization}
\end{align}

The model described above is summarized in \cref{fig:overview}. We now proceed to describe how each component is implemented in \cref{sec:implementation}. 

\section{\textsc{RelDetect} - Model Implementation} \label{sec:implementation}
In this section, we describe our implementation of the three components of our model: coarse detector, the relationship prediction network and the fine-grained classifier. 

\subsection{Region Proposal Network and Coarse Classifier} \label{sec:rpn}
Our Region Proposal Network \texttt{RPN} is implemented as a neural network that predicts bounding boxes $S$ given the image $I$. We use the region proposal network $f_{\rm RPN}$ from an off-the-shelf object detector, Faster RCNN \cite{ren2016faster}. Each spatial description $S_i \in \R^4$ is a 4-tuple containing the height, width and the center $x, y$ coordinates of the bounding box. Note that there is no uncertainty involved in the spatial description, hence the probability model $p(S | I)$ can be seen as a dirac delta defined as:
\begin{equation}
p(S | I) = \delta_{f_{\rm RPN}(I)}
\end{equation}

The coarse classifier is then implemented as an object classification head $f_{\rm coarse}$ on top of image features extracted from each predicted $S_i$. The feature extraction is performed using RoI Align \cite{he2017mask} on top of features extracted by a deep network from the image $I$. The probability model is given by:
\begin{equation}
p(\tilde C_i | S_i, I) = f_{\rm coarse}(\textsc{Roi-Align}(S_i, I))
\end{equation}
\subsection{Relationship Prediction Network} \label{sec:relpn}
Our Relationship Prediction Network \texttt{REL} is implemented as a network that takes as input the predicted bounding boxes $S_1, \ldots, S_N$ and the predicted classes $C_1, \ldots, C_N$ and outputs the relationship probabilities for all $N^2$ pairs of objects. Note that since the number of relationships to be predicted is quadratic in $N$, a non-maximum suppression step is applied to the bounding boxes as a preprocessing step to make computation feasible. We follow the architecture design in \cite{zellers2018neural,tang2020unbiased} and set up \texttt{REL} as a bidirectional LSTM followed by a MLP. 

An object representation $\mathbf{o}_i$ is computed for each object $i$ by concatenating an embedding of the predicted coarse class, the RoI-Align feature, and an spatial embedding $f_{\rm SP}$ of the predicted bounding box $S_i$:
\begin{equation}
\mathbf{o}_i = [\mathbf{W}_1 \mathbf{\tilde c}_i; \textsc{Roi-Align}(S_i, I); f_{\rm SP}(S_i)]
\end{equation}
In the above, $\mathbf{W}_1$ is a learnt embedding and $\mathbf{\tilde c}_i \in \{0, 1\}^{\tilde K}$ is a one-hot vector representing the predicted coarse class. 

A global contextualized embedding $\mathbf{e}_i$ is then computed for each object using a bidirectional LSTM on the object embedding sequence $\{\mathbf{o}_i\}$:
\begin{equation}
\mathbf{e}_i = {\rm biLSTM}([\mathbf{o}_1, \mathbf{o}_2, \ldots, \mathbf{o}_N])
\end{equation}

Finally, for every pair $\mathbf{e}_i, \mathbf{e}_j$, the relationship probability model is given by a network $f_{\rm REL}$ that additionally takes as input the RoI-Align feature computed from the union box corresponding to $S_i, S_j$ denoted as $S_i \cup S_j$:
\begin{equation}
p(R_{i, j} | \tilde C, S, I) = f_{\rm REL}(\mathbf{e}_i, \mathbf{e}_j, \textsc{Roi-Align}(S_i \cup S_j, I))
\end{equation}
\subsection{Knowledge Base Model} \label{sec:kb}
Our knowledge base $B$ consists of relationship tuples linking the base classes to the fine-grained classes via relationships. These tuples are extracted from common-sense knowledge sources, \eg Wikipedia, and take the form ``an object $i$ belonging to \texttt{coarse-class} $\tilde c_2$ should be classified as \texttt{fine-grained-class} $c$ when there is a relation \texttt{relationship} $r$ between object $i$ and object $j$ which belongs to \texttt{coarse-class} $\tilde c_2$''. The collection of such tuples forms our knowledge base $B$: 
\begin{equation}
B = \{(c_i, \tilde c_{i, 1}, \tilde c_{i, 2}, r_i)\}
\end{equation}

Now given $B$ and a predicted bounding box $i$, we search for possible knowledge-base matches among the neighboring boxes of $i$ \ie $N_i$. This gives the final probability model for the fine-grained class prediction:
\begin{align}
p(C_i | R_{i, {\rm NB}_i}, \tilde C_{{\rm NB}_i}) = \OR_{j \in {\rm NB}_i} \mathbbm{1}[(C_i, \tilde C_i, \tilde C_j, R_{ij}) \in B] \label{eq:kb}
\end{align}
In the above the $\OR$ operator evaluates to $1$ if any of the operands is $1$, and evaluates to $0$ otherwise. 

During inference, given an image, we first of all use the neural network \texttt{RPN} for predicting a set of  bounding boxes in the image, which is then refined by applying non-maximum suppression. The obtained boxes are then classified into coarse categories by the object classification head $f_{\rm coarse}$. The relationship prediction network \texttt{REL} is then used to predict all relationships between the obtained bounding boxes. Finally, the knowledge base model \cref{eq:kb} is applied using the predicted coarse classes and relationships to obtain the fine-grained predictions. This completes the description of our model  \textsc{RelDetect}. We now proceed to describe the experimental protocols and results in \cref{sec:expt}. 

\begin{table*}[t!]
  \centering
   \begin{tabular}{cC{3.25cm}C{2cm}C{1cm}C{2.5cm}C{1cm}C{2.5cm}}
   \toprule
 Method & DeepRooms Labels Used & Relationship Annotations Used & \multicolumn{2}{c}{mAP DeepRooms} & \multicolumn{2}{c}{mAP LivingRoom} \\
\cmidrule[1pt]{4-5} \cmidrule[1pt]{6-7} 
& &  & Coarse & Fine-Grained & Coarse & Fine-Grained \\
\midrule
\textsc{RelDetect} 	& Coarse						& \texttt{AUTO} ($0.2\%$) 	& 95.59 			& 75.46 			& 71.64 			& 52.70 \\
\textsc{RelDetect} 	& Coarse						& \texttt{AUTO} ($2\%$)  	& 95.97 			& \textbf{82.17} 	& 72.18 			& 55.33 \\
\textsc{RelDetect} 	& Coarse						& \texttt{AUTO} ($4\%$)  	& 96.00 			& 79.78 			& 72.26 			& 55.36 \\
\textsc{RelDetect} 	& Coarse						& \texttt{AUTO} ($10\%$)  	& \textbf{97.55} 	& 79.36			 	& \textbf{75.16} 	& \textbf{56.45} \\
\textsc{RelDetect} 	& Coarse						& \texttt{AUTO} ($40\%$)  	& 96.26 			& 75.21 			& 74.30 			& 53.82 \\
\textsc{RelDetect} 	& Coarse						& \texttt{AUTO} ($80\%$)  	& 95.90 			& 70.16 			& 73.46 			& 50.89 \\
\textsc{RelDetect} 	& Coarse						& \texttt{AUTO} ($100\%$) 	& 95.53 			& 70.48 			& 73.30 			& 49.94 \\
Fine-Tuning 		& Fine-Grained ($0.2\%$) 	& $-$						 		& 78.56					& 78.28 			& 49.86					& 46.86 \\
\midrule
Fine-Tuning 		& Fine-Grained ($100\%$) 	& $-$						 		& 89.85					& 94.89 			& 75.29					& 68.97 \\
 \bottomrule
   \end{tabular}
  \caption{Experimental Results: Comparing Few-Shot Capabilities. The metric reported is mAP@50 over the Coarse or Fine-Grained classes. Detailed per-class AP scores can be found in the Appendix. No labels from the LivingRoom dataset were used for training. We use Faster RCNN as the detector for coarse classes. For any row, a dashed ($-$) column indicates an invalid setting \eg, the fine-tuning baselines do not need any relationship annotations for training. See \cref{sec:fsod} for details and experimental protocol.}
  \label{tab:mainfsod}
\end{table*}

\section{Experiments} \label{sec:expt}
In this section, we describe our experimental studies. We first describe the datasets that we use for evaluation, and then detail our experimental protocols and results.

\textbf{Datasets} We show experimental results on two datasets curated for the task of furniture detection: (1) DeepRooms \cite{gadde2021detail}, having $48,474$ images with $169,877$ bounding box annotations and (2) LivingRoom, having $1506$ images with $8,193$ bounding box annotations. 
Each dataset consists of images of living rooms with furniture items (objects) labelled with the 8 fine-grained classes \{\texttt{chair}, \texttt{sofacouch}, \texttt{coffee-table}, \texttt{end-table}, \texttt{console-table}, \texttt{table-lamp}, \texttt{floor-lamp}, \texttt{rug}\} corresponding to the 5 coarse classes \{\texttt{chair}, \texttt{sofacouch}, \texttt{table}, \texttt{lamp}, \texttt{rug}\}. There are no relationship labels available in the datasets. For the purposes of our experiments, we will outline our \textsc{Auto} protocol to obtain the relationship labels. 

We clarify two important aspects about our experiments on these datasets:
\begin{itemize}
\item The training split of the DeepRooms dataset is used for learning our detector and we use the testing split for demonstrating our first main result in \cref{sec:fsod}: Detection using relationships provides comparable extremely few-shot performance than fine-tuning based methods \cite{wang2020frustratingly}.
\item The LivingRoom dataset contains more in-the-wild images than DeepRooms. We use it only for testing transfer performance, and  demonstrating our second main result in \cref{sec:transfer}: Detection using relationships provides stronger zero-shot domain transfer performance than fine-tuning based methods when very few fine-grained labels are available.
\end{itemize}
 
\textbf{Fine-Tuning Baselines} We compare \textsc{RelDetect} to fine-tuning baselines in \cref{tab:mainfsod,tab:maintransfer}. The detectors used are FasterRCNN \cite{ren2016faster} and YOLOv3 \cite{redmon2018yolov3}. In all the fine-tuning baselines, the coarse detector was first of all pre-trained on a class converted version of DeepRooms, where all the fine-grained labels were converted to their corresponding coarse labels. The the detectors were then fine-tuned on fine-grained labels (hence the name Fine-Tuning baseline). In each table, the dataset and the size of these fine-grained labels is mentioned on each fine-tuning row.

\subsection{Few-Shot Detection using Relationships} \label{sec:fsod}
In order to demonstrate the few-shot detection capabilities of \textsc{RelDetect} in \cref{tab:mainfsod}, we follow an experimental protocol of using a fraction of the fine-grained labels available in our training dataset, DeepRooms. 

\begin{table*}
  \centering
   \begin{tabular}{cC{3.25cm}C{2cm}C{3.25cm}C{2cm}C{1cm}C{1cm}C{1cm}}
   \toprule
 Method & DeepRooms Labels Used & Relationship Annotations Used & LivingRoom Labels Used & Coarse Detector & \multicolumn{2}{c}{mAP LivingRoom} \\
\cmidrule[1pt]{6-7}
 & &  &  &  & Coarse & Fine-Grained \\
\midrule
\textsc{RelDetect} 	& Coarse						& \texttt{AUTO} ($0.2\%$)	& None & Faster RCNN 	& 71.64 			& \textbf{52.70} \\
Fine-Tuning & Fine-Grained ($0.2\%$) & $-$ & None & Faster RCNN & 49.86 & 46.86 \\
\midrule
Fine-Tuning & Fine-Grained ($100\%$) & $-$ & None & Faster RCNN & 89.85 & 68.97 \\
Fine-Tuning & Coarse (100\%) & $-$ & Coarse (100\%) & YOLOv3 & 80.03 & $-$ \\
Fine-Tuning & Coarse (100\%) & $-$ & Fine-Grained ($11\%$) & YOLOv3 & 67.56 & 60.34 \\
Fine-Tuning & Fine-Grained ($100\%$) & $-$ & None & YOLOv3  &  70.77 & 62.96 \\
Fine-Tuning & Fine-Grained ($100\%$) & $-$ & Fine-Grained ($100\%$) & YOLOv3 & 79.62 & 72.56\\
 \bottomrule
   \end{tabular}
  \caption{Experimental Results: Comparing Zero-Shot Domain Transfer Capabilities. The metric reported is mAP@50 over the Coarse or Fine-Grained classes. Detailed per-class AP scores can be found in the Appendix. For any row, a dashed ($-$) column indicates an invalid setting \eg, a fine-tuning baseline trained with only coarse labels cannot predict fine-grained categories. See \cref{sec:transfer} for details.}
  \label{tab:maintransfer}
\end{table*}

\textbf{Experimental Protocol} More specifically, the experimental protocol consists of two steps.
\begin{enumerate}
\item To mimic a coarse-grained object detector trained on a large dataset, we train the methods on coarse-grained labels \{\texttt{chair}, \texttt{sofacouch}, \texttt{table}, \texttt{lamp}, \texttt{rug}\} on the entire dataset. 
\item Then, to mimic availibility of a small amount of fine-grained task specific labels,  \{\texttt{chair}, \texttt{sofacouch}, \texttt{coffee-table}, \texttt{end-table}, \texttt{console-table}, \texttt{table-lamp}, \texttt{floor-lamp}, \texttt{rug}\}, we only allow the methods to access $x\%$ of the images in the entire dataset, annotated with fine-grained object labels. $x$ is varied from $0.2$ to $100$. 
\end{enumerate}
\textbf{Auto Relationship Annotation} Since \textsc{RelDetect} needs relationship annotations between object boxes labelled with the coarse classes, we propose a method for automatic relationship annotation (\texttt{AUTO}) using the only $x\%$ fine-grained labels (as described above). In \texttt{Auto}, we find object pairs in the image that satisfy the knowledge base, \ie for a tuple $\{(c, \tilde c_1, \tilde c_2, r)\}$, we find all boxes $i, j$ such that the coarse class of object $i$ is $\tilde c_1$ and that of object $j$ is $\tilde c_2$, and the  fine-grained class of object $i$ is $\tilde c$. Among such pairs $(i, j)$, we retain only those pairs where the centers of the predicted bounding boxes are at most a fixed distance $\tau$ from each other \footnote{We find that this fixed-distance based filtering scheme works reasonably for our datasets. Dynamic neighborhood generation schemes might perform better for datasets having a large variation in object sizes.}. These retained pairs are labelled with the corresponding relationship $r$. These annotated relationships are used to train the relationship prediction network \texttt{REL} from \cref{sec:relpn}, and this process is labelled \texttt{AUTO} $x\%$ in \cref{tab:mainfsod}. 

With the above protocol, we compare \textsc{RelDetect} to a strong baseline which is trained with access to all ($100\%$) of the fine-grained and coarse-grained labels in DeepRooms. The results are presented in \cref{tab:mainfsod}. In order to keep the comparison fair across datasets of widely varying sizes, we have fixed the total number of iterations that each method is trained with to $10,000$ iterations. This in turn also ensures that each row is trained for the same amount of GPU time.

\textbf{Result Analysis} Firstly, we observe that training using relationships inferred from fine-grained annotations on $0.2\%$ of DeepRooms images leads to a performance which is comparable to training an object detector with the same data, \ie $75.46$ mAP vs $78.28$ mAP. This demonstrates that learning with relationships approximately retains performance in low-data regimes. As the fraction of available fine-grained annotations ($x$) is increased, we see that the performance on DeepRooms gradually increases, peaking at $x = 2$, where the performance is within $12$ mAP points of the baseline trained with $100\%$ of the data ($82.17$ mAP vs $94.89$ mAP). While this gap might seem significant, note that \textsc{RelDetect} is trained using no fine-grained bounding boxes, while the Fine-Tuning baseline is trained with fine-grained bounding boxes on $50\times$ more image data. This demonstrates that learning with relationships provides a good trade-off between obtaining fine-grained annotations and performance. We observe that increasing $x$ further deteriorates the performance. We believe this might be due to the inaccuracies in the \texttt{AUTO} annotation rule which might negate the benefits of learning with relationships. On the other hand, we note that the fine-tuning baseline trained on the full dataset is able to obtain the best Fine-Grained performance on DeepRooms, \ie $94.89$ mAP, reassuring us that training a standard detector for a long time with task-specific labels does provide good performance on the task. 

\begin{figure*}
  \centering
  \includegraphics[width=\textwidth]{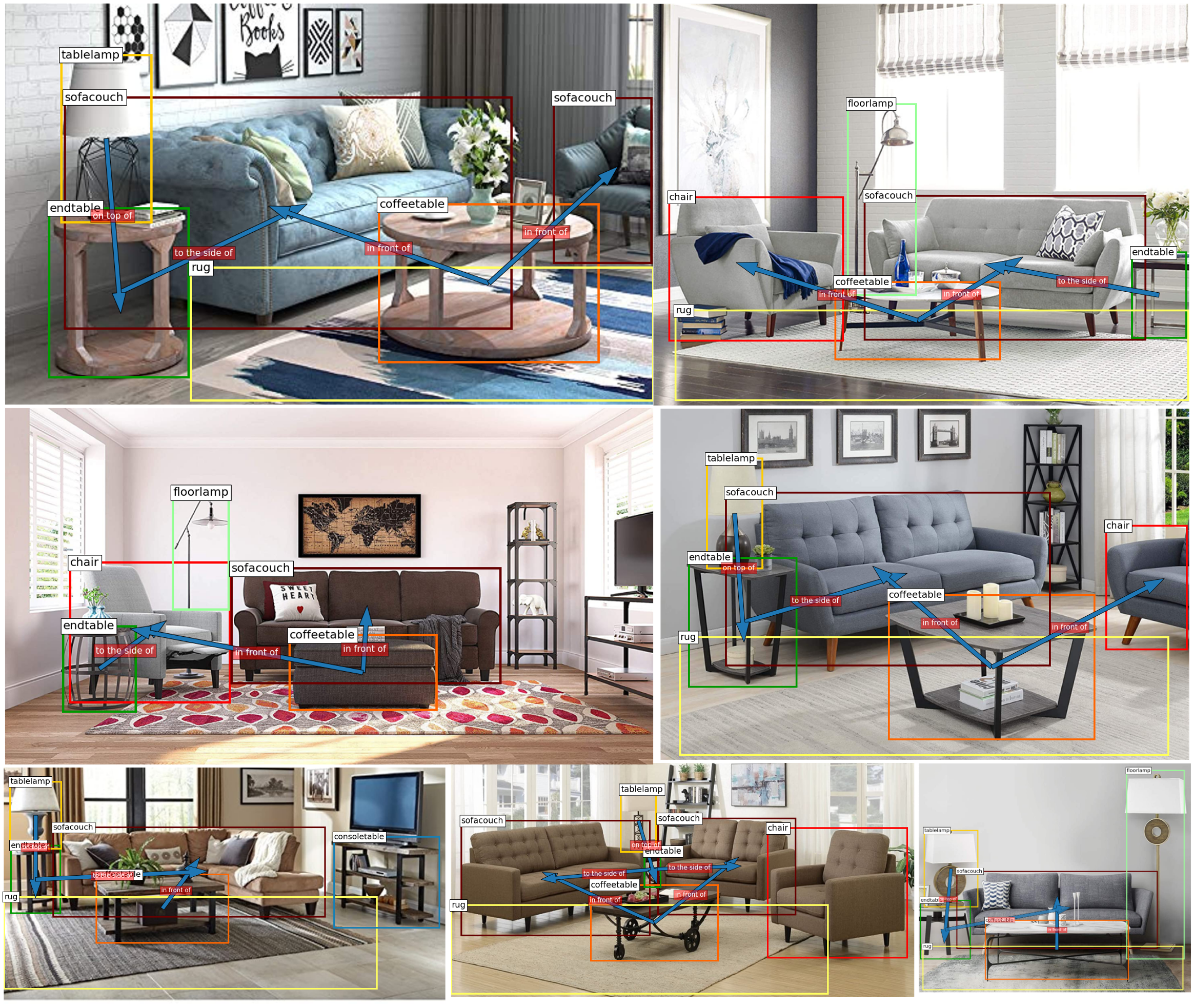}
  \caption{Qualitative Results from our method \textsc{RelDetect}. Considering the top-left image: The bounding boxes show the fine-grained labels output from \textsc{RelDetect}. We draw an arrow with the label $r_{i, j}$ between the centers of two bounding boxes $i, j$ when the probability of corresponding relationship $r_{i, j}$ is above a threshold. In this example, the base categories ``lamp'' (yellow) and ``table'' (green) were predicted, and the relationship ``on top of'' was predicted, leading to a fine-grained label of ``table-lamp''. Best viewed in color.}
  \label{fig:qualpos}
\end{figure*}

\subsection{Zero-Shot Domain Transfer using Relationships} \label{sec:transfer}
In this section we compare the first \textsc{RelDetect} row in \cref{tab:mainfsod} to fine-tuning baselines in order to understand the benefits of learning with relationships when we have no access to labels from a target domain. 

\textbf{Experimental Protocol} For our method \textsc{RelDetect}, the protocol remains the same as \cref{sec:fsod}. For the finetuning baselines, we allow access to $y \%$ ($y \in \{0, 11, 100\})$\footnote{$11\%$ corresponds to $1000$ bounding box annotations in LivingRoom} of fine-grained labels from the target domain, which is taken to be the LivingRoom dataset. We then report the mAP on the target domain in \cref{tab:maintransfer}. 

\textbf{Result Analysis} We firstly observe (from \cref{tab:mainfsod}) that the source domain (DeepRooms) and target domain (LivingRoom) have a large distribution shift, as all methods drop around $20$ to $30$ mAP points from DeepRooms to LivingRoom. This also suggests that detection on LivingRoom is a harder task. Secondly, we observe that \textsc{RelDetect} is able to utilize relationships in a low-data regime ($0.2\%$ fine-grained labels) to improve performance on LivingRoom by $+5.68$ mAP considering a baseline trained with the same number fine-grained annotations, \ie $52.70$ mAP vs $47.02$ mAP. This demonstrates that relationships are good for learning transferrable representations, which is the second main experimental insight from this section. Finally, we observe from the last few rows of \cref{tab:maintransfer} that training on even a small fraction ($11\%$) of fine-grained labels from the target domain leads to a large increase in performance, reassuring us that training with good task-specific labels provides the best performance. Having analysed \textsc{RelDetect} quantitatively, we now show a qualitative evaluation in \cref{fig:qualpos}.

\section{Conclusion, Shortcomings and Future Work}

In this work, we proposed \textsc{RelDetect}, a relationship based probabilisitic model for fine-grained object detection. \textsc{RelDetect} utilizes relationships between coarse-grained object classes and an external knowledge base to predict fine-grained classes. We demonstrated via experiments that such learning from relationships enables us to obtain reasonable performance in limited data regimes where we have access to a small amount of fine-grained training data (few-shot), and superior performance where we have access to no training data from a target domain (zero-shot transfer). 

In this work, we limit ourselves to datasets where there is a tight coupling between the coarse-grained classes and the fine-grained classes via suitable relationships. Our method is intended for settings where we can obtain a set of common sense rules (\ie, the knowledge-base $B$ in our case) where one can infer a fine-grained object category by looking at relationships between objects belonging to coarse categories in the scene. While this hierarchy is very natural in many fine-grained object detection tasks, there are tasks where such a hierarchy is not readily available, \eg, detecting and categorizing fine-grained breeds of dogs. However, recent works in zero-shot learning (ZSL) 
have developed methods to identify a suitable hierarchy from large text and image corpora. More specifically, we have the hierarchy coarse-grained $\to$ relationships $\to$ fine-grained, whereas the ZSL literature works with base classes $\to$ class attributes $\to$ novel-classes. Such attributes would lead to a natural generalization of our relationship-based detection framework. Our relationship prediction network would then generalize to an attribute prediction network, which takes as input one or more coarse-objects, and predicts an attribute. 

While we show improvements with our noisy \texttt{AUTO} annotation pipeline, it can often lead to inaccurate relationship labels depending on the distance threshold $\tau$. A better method for relationship annotation, \eg selecting $\tau$ dynamically per image, could lead to an improvement in  \textsc{RelDetect} performance.

{\small
\bibliographystyle{ieee_fullname}
\bibliography{RelDet_WACV23.bib}
}

\end{document}